\documentclass{article}

\usepackage{microtype}

\usepackage{graphicx}
\usepackage{subcaption} 

\usepackage{booktabs} 

\usepackage{enumitem}
\usepackage{tabularx}
\usepackage{multicol}
\usepackage{makecell}

\usepackage{makecell}

\usepackage{hyperref}
\usepackage{cleveref} 

\usepackage{natbib}

\usepackage[accepted]{icml2019}

\icmltitlerunning{Image Matters: Detecting Offensive and Non-Compliant Images}

\begin{document}

\twocolumn[
\icmltitle{Image Matters: Scalable Detection of Offensive and Non-compliant Content / Logo in Product Images}



\icmlsetsymbol{equal}{*}

\begin{icmlauthorlist}
\icmlauthor{Shreyansh Gandhi}{wmt}
\icmlauthor{Samrat Kokkula}{wmt}
\icmlauthor{Abon Chaudhuri}{wmt}
\icmlauthor{Alessandro Magnani}{wmt}
\icmlauthor{Theban Stanley}{wmt}
\icmlauthor{Behzad Ahmadi}{wmt}
\icmlauthor{Venkatesh Kandaswamy}{wmt}
\icmlauthor{Omer Ovenc}{wmt}
\icmlauthor{Shie Mannor}{ed}
\end{icmlauthorlist}

\icmlaffiliation{wmt}{Walmart Labs, Sunnyvale, California, USA}
\icmlaffiliation{ed}{Department of Electrical Engineering, Technion, Israel}

\icmlcorrespondingauthor{Shreyansh Gandhi}{sgandhi@walmartlabs.com}


\icmlkeywords{e-commerce, machine learning, deep learning, computer vision, image understanding, offensive images, non-compliant images, content moderation, image classification, object detection, pattern recognition, logo detection, synthetic data generation}
\vskip 0.1in
]



\printAffiliationsAndNotice{}  

\begin{abstract}
   In e-commerce, product content, especially product images have a significant influence on a customer's journey from product discovery to evaluation and finally, purchase decision. Since many e-commerce retailers sell items from other third-party marketplace sellers besides their own, the content published by both internal and external content creators needs to be monitored and enriched, wherever possible. Despite guidelines and warnings, product listings that contain offensive and non-compliant images continue to enter catalogs. Offensive and non-compliant content can include a wide range of objects, logos, and banners conveying violent, sexually explicit, racist, or promotional messages. Such images can severely damage the customer experience, lead to legal issues, and erode the company brand. In this paper, we present a computer vision driven offensive and non-compliant image detection system for extremely large image datasets. This paper delves into the unique challenges of applying deep learning to real-world product image data from retail world. We demonstrate how we resolve a number of technical challenges such as lack of training data, severe class imbalance, fine-grained class definitions etc. using a number of practical yet unique technical strategies. Our system combines state-of-the-art image classification and object detection techniques with budgeted crowdsourcing to develop a solution customized for a massive, diverse, and constantly evolving product catalog.
\end{abstract}

\setlist{nolistsep}

 \section{Introduction}
\label{sec:intro}
By nature, humans are visual learners. A single snapshot of a product provides more information about the product than a wall of text. According to a research from $``$Nielson Norman$"$\cite{reading_numbers},  only 16$\%$ of the readers actually read word-for-word and 79$\%$ only gloss over the highlights. In e-commerce, good quality images help customers understand the product better, motivate them to read about it, and build customers' trust in the product quality. This eventually increases the chance of actual purchase by the customer. 

Despite the well-known importance of images, e-commerce retailers, especially the ones who allow marketplace items from 3$^{rd}$ party sellers, struggle to control image quality. Both external and internal content creators are expected to meet the retailer's Trust $\&$ Safety guidelines. However, these guidelines constantly change and expand, which makes it incredibly difficult for e-commerce retailers to ensure that external content providers are complying with  guidelines. This is why e-commerce retailers are interested to automate the process of content validation and filtering using computer vision and related technology.


\begin{figure}[ht]
\begin{center}
		\includegraphics[width=\columnwidth]{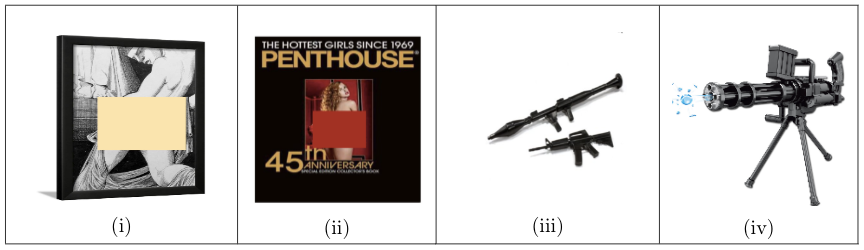}
		\caption{Examples of offensive/non-compliant content : i) nudity ii) sexually explicit iii) assault rifle iv) toy resembling assault rifle)}
		\label{fig:examples_offensive_images}
\end{center}
\end{figure}


\begin{figure}[ht]
\begin{center}
  \includegraphics[width=\columnwidth]{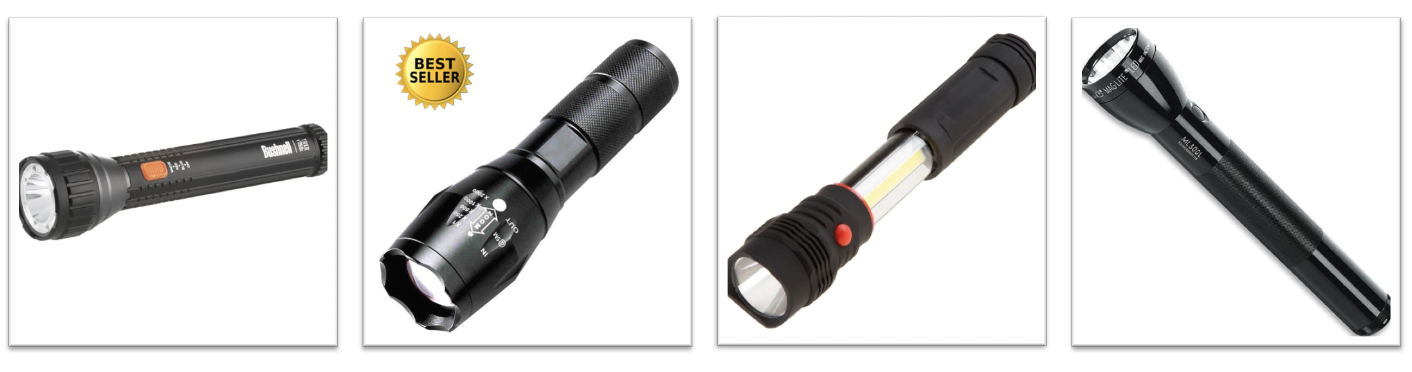}
\caption{Which product would you choose? Promotional logos such as $``$best seller" is considered non-compliant.}
\label{fig:_Which_product_would_you_choose}   
\end{center}
\end{figure}

\begin{figure}[ht]
\begin{center}
  \includegraphics[width=\columnwidth]{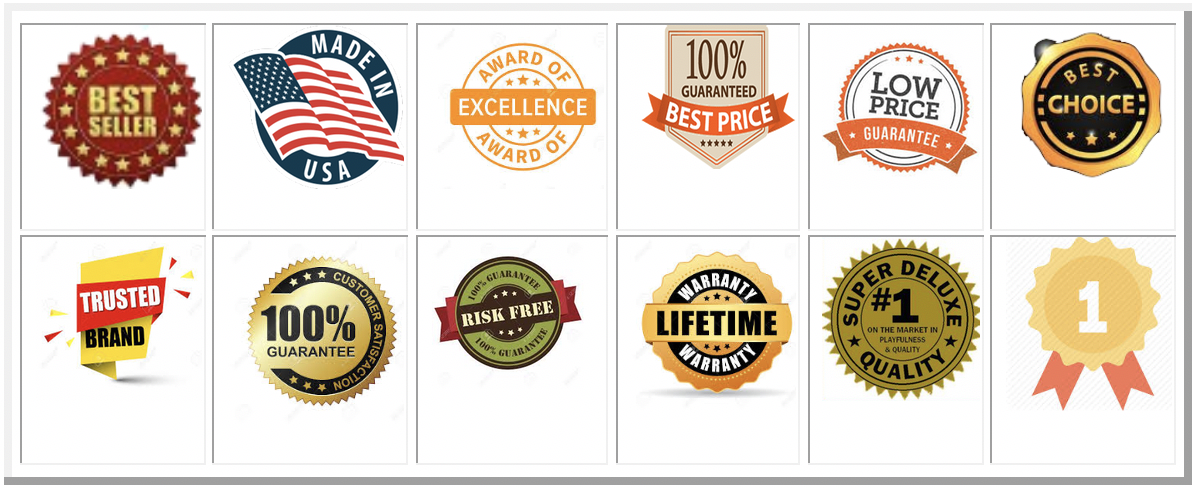}
\caption{Examples of marketing badges (includes award badges, the seal of excellence, best-price guarantee, lowest price, made in USA, manufactured/assembled in USA, etc.}
\label{fig:_Examples_of_Marketing_Badges}  
\end{center}
\vskip -0.2in
\end{figure}

Trust $\&$ Safety guidelines usually encompass following broad categories:
\begin{enumerate}
    \item \textit{\underline{Offensive Images}}: Figure~\ref{fig:examples_offensive_images} shows various types of offensive images. The examples include images that have nudity, sexually explicit content, abusive text, objects used to promote violence, and racially inappropriate content. 
    \item \textit{\underline{Non-compliant Images}}: Most e-commerce retailers have published compliance guidelines on what can be sold on their platform.  Figure~\ref{fig:examples_offensive_images} [iii] and [iv] shows images of products that are non-compliant such as assault rifles and a toy that resembles assault-style rifle. 
    \item \textit{\underline{Logos and Badges}} : A wide range of logos and banners are considered non-compliant too. In Figure~\ref{fig:_Which_product_would_you_choose}, the image located second from the left uses a self proclaimed marketing logo to lure the customer to click on it. This is a common malpractice and such logos are considered non-compliant. Other non-compliant logo types include competitors' logos, inaccurate manufacturing country logos (e.g., Made in USA logo), and many others (as seen in Figure~\ref{fig:_Examples_of_Marketing_Badges}).
\end{enumerate}
 
Traditionally, the retailers try to address this problem either by displaying a disclaimer on the website that the displayed content is not owned by the retailer, or by allowing the customers to report non-compliant content so that they can be filtered by a human workforce. Unfortunately, these options do not protect the customer from having an unpleasant experience from seeing such images. Also, the disclaimer often goes unnoticed and the retailers brand value is tarnished. Most importantly, these solutions do not scale.

In this paper, we present a computer vision based system that automates the image detection and filtering process for an extremely large catalog of images, and helps the retailer enforce its compliance terms and conditions. We discuss in detail how we blend human expertise with state-of-the-art deep learning models to overcome a number of data and system level technical challenges outlined in Section~\ref{sec:challenges}.

The core learnings from this system can be utilized by any system that serves image or other visual content to human users on the web such as social media feeds, ads platforms, etc.
 \section{Technical Challenges}
\label{sec:challenges}
The proposed system is designed to address a number of data and system-level challenges as described below:

\begin{figure}[ht]
\includegraphics[width=\columnwidth]{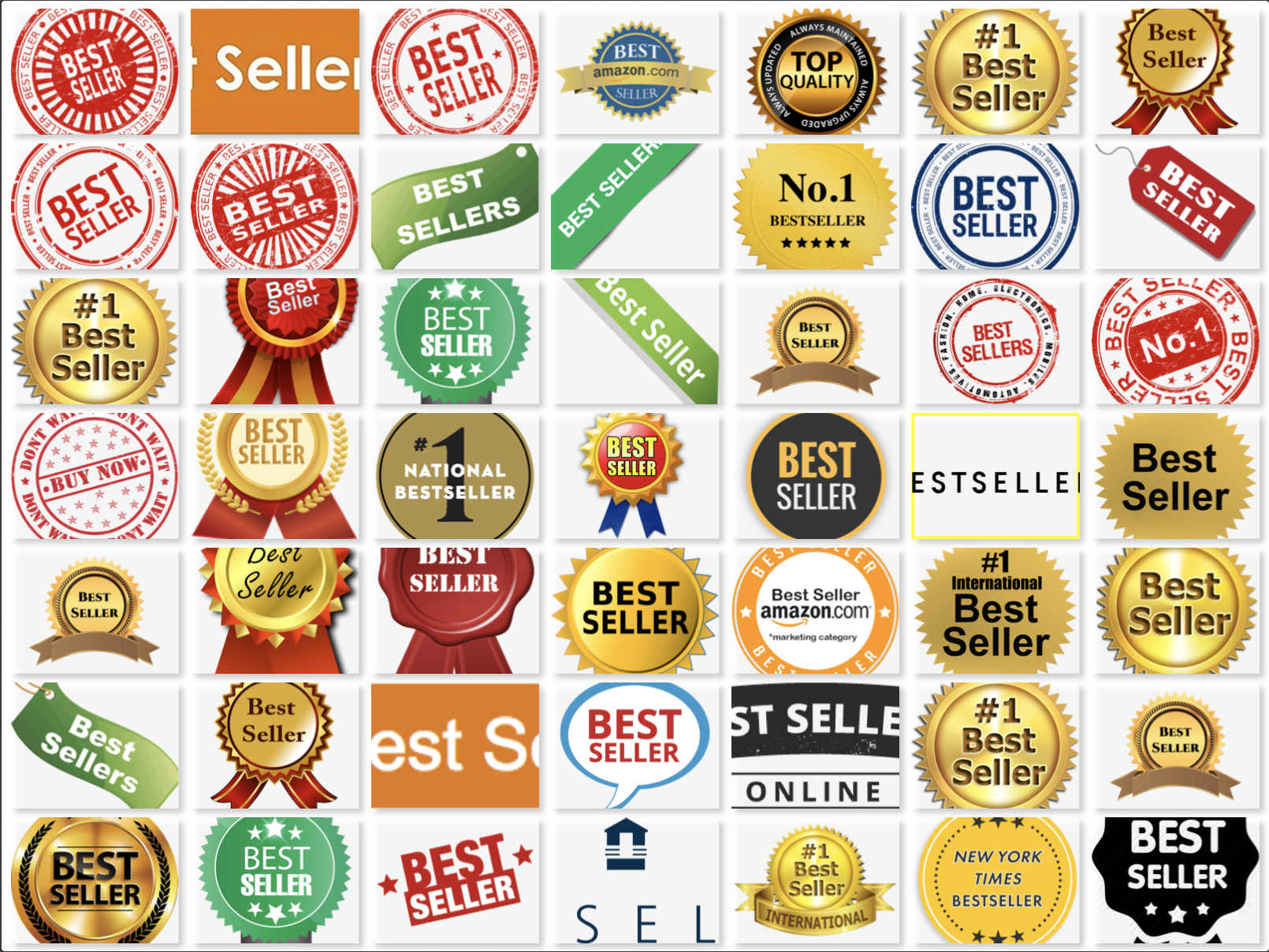}
\caption{Challenge 3: Non-compliant category (e.g. Best Seller Badges) has various forms in which it can appear on images.}
\label{fig:_Examples_of_Best_Seller_Badges} 
\vskip -0.1in
\end{figure}


\begin{enumerate}
    \item \textbf{Lack of Usable Training Data} : Most non-compliant images are hard to find. In most cases, the first example is discovered and reported by a customer. Even if we collect similar images from various sources on the internet, it is tens of labeled data points at best. Manual tagging is prohibitively expensive because the crowd needs to review thousands of images to find one true non-compliant example.
    \item \textbf{Scale and Variation in Catalog}: E-commerce catalogs of large retailers have hundreds of millions of products, across tens of thousands of product categories. Additionally, the non-compliant content of any given type can appear across several, if not all, product categories. For example, the $``$best-price" logo can appear on images of products from any category. Moreover, the catalog data keeps changing. Creating a big enough training set that is a true representative of the catalog is difficult and expensive.
    \item \textbf{Variety of Defining Examples}: A single non-compliant type can appear in multiple forms (e.g., a best seller badge) (Figure~\ref{fig:_Examples_of_Best_Seller_Badges}). We need to ensure that our models are generalized enough to capture various forms of infractions for a single use case.
    \item \textbf{Custom and Fine-Grained Class Definitions} : The non-compliance guidelines often apply to a certain form or variation of an object. For example, most e-commerce websites allow hunting rifles but not assault rifles. From a machine learning point of view, differentiating between assault rifle and hunting rifle images falls into the category of fine-grained classification which is challenging. Similarly, the image of a person wearing a swimwear is acceptable, but a picture of a nude person which is visually close to the former is not acceptable. This also means standard object detectors that detect guns or people would not suffice to solve our problem. An even more difficult manifestation of the problem is the case where certain images (such as a swimwear) are deemed offensive because of the pose or expression of the person, but other images featuring the same person are considered acceptable. 
    \item \textbf{Constraints on using text}: Even though each product comes with large amount of rich textual data, they are not easy to use for this problem. It is quite common for a compliant product to have a non-compliant image (e.g., a music CD with a nude photo on the CD) or vice versa. A title-based detector would fail to capture such an example. Alternately, optical character recognition (OCR) can be used to extract non-compliant text from the images alone. However, OCR works only if the image meets certain conditions. Also, OCR cannot capture a wide range of problems, such as nudity or an assault rifle, where there is no text on the image.
\end{enumerate}

 \section{Related Work}
\label{sec:related_work}

The importance of images in e-commerce is well studied. Online shoppers often use images as the first level of information. Also, item popularity highly depends on the image quality~\cite{qualitypopularity}. \cite{image_ecom_14} provides deeper understanding of the roles images play in e-commerce and shows evidence that better images can lead to an increase of buyers' attention, trust, and conversion rates.

Image classification models based on skin detection techniques~\cite{arentz2004classifying} have been proposed for nudity detection. Skin regions are detected based on color, texture, contour, and shape information features.
\cite{zheng2004blocking} uses maximum entropy distribution to detect the skin regions in the image.


Traditionally, logo/badge recognition has been addressed by keypoint-based detectors and descriptors \cite{kalantidis2011scalable,romberg2013bundle,joly2009logo}, feature detection (using SIFT, SURF, BRIEF, ORB), and feature matching (using Brute-Force, FLANN matcher) \cite{feature_matching} and classical template matching \cite{template_matching}. From our experience, these techniques do not work well for a catalog that contains millions of products. A few deep learning based logo detection
models have been reported recently \cite{bianco2017deep,su2017deep,eggert2015benefit,iandola2015deeplogo}. All of these techniques are tested on publicly available brand-logo datasets like BelgaLogos \cite{joly2009logo} or FlickrLogos-32 dataset \cite{romberg2011scalable}. 


Recent advances in deep learning have brought neural nets to the forefront of image classification. A number of deep learning architectures such as Alexnet~\cite{alexnet}, VGG net~\cite{simonyan2014very}, residual network~\cite{he2016deep}, Inception~\cite{inceptionv1,inceptionv2}, and Nasnet~\cite{nasnet} have been proposed. In this paper, we experimented with Resnet and Inception architectures that were pre-trained on an  Imagenet~\cite{imagenet_cvpr09} dataset, and then retrained on our images.

Object detection deals with detecting instances of semantic objects of a certain class and identifying the location of them. Some well-known object detectors are SSD\cite{ssd}, Region-based object detection\cite{girshick2016region}, YOLO\cite{YOLO}, R-FCN\cite{dai2016r}, and Faster R-CNN\cite{ren2015faster}.

Generating synthetic training data allows for expanding plausibly ground-truth annotations without the need for exhaustive manual labelling. This strategy has been shown to be effective for training large CNN models, especially when sufficient training data is not available \cite{dosovitskiy2015flownet,eggert2015benefit}. 

 \section{Proposed Detection Framework}
\label{sec:framework}
\begin{figure}
\begin{center}
	\includegraphics[width=\columnwidth]{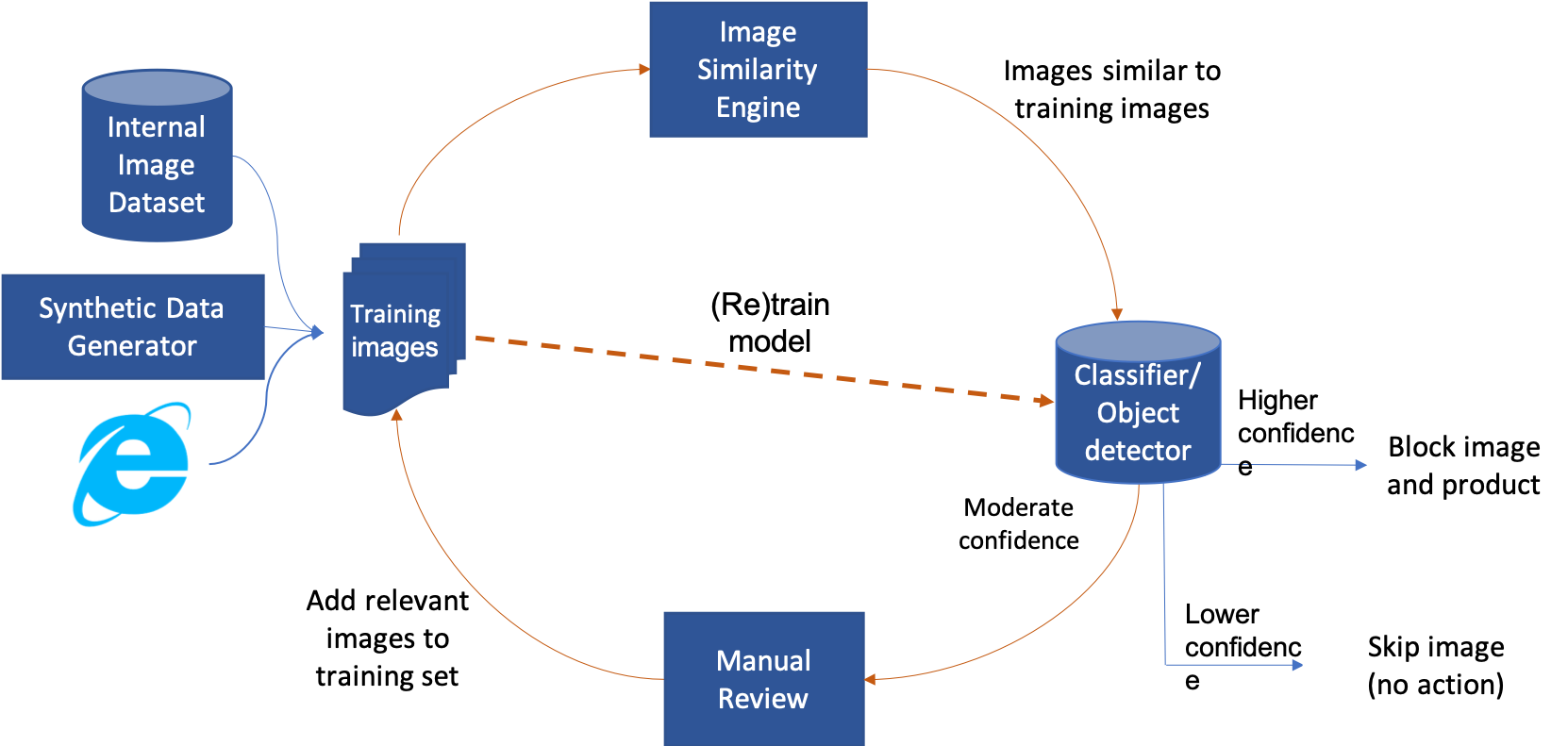}
	\caption{Proposed offensive and non-compliant image detection framework}
	\label{fig:Learning_Cycle}
\end{center}
\vskip -0.2in
\end{figure}

In this paper, we propose a computer vision powered framework, as outlined in Figure~\ref{fig:Learning_Cycle}, for sparsely occurring content detection from images. In order to address the extreme scale, diversity, and dynamism of our dataset, we deviate from conventional approaches and innovate in a couple of ways.
\begin{enumerate}
    \item \textbf{Iterative Training:} Unlike well-posed machine learning problems, we often start with a handful examples of offensive or non-compliant images. Hence, we collect data from various auxiliary sources and iterate a few times, as described in Section~\ref{sec:training_data} to build training data.
    \item \textbf{Transfer Learning:} It is impossible to train a neural net from scratch with the amount of data we have. Hence, we leverage pre-trained networks and fine tune them with small but carefully crafted training data. (Section~\ref{sec:approaches})
    \item \textbf{Multi-stage Inference:} In order to scale, we propose to combine faster and lightweight classifiers with slower and deeper object detection networks. (Section~\ref{sec:approaches})
\end{enumerate}


\subsection{Training Data Augmentation}
\label{sec:training_data}




Standard image data augmentation techniques such as translation, flip, rotation, color/contrast adjustment and noise incorporation are not sufficient for our application because we often start with a minuscule number of images. We use the above mentioned controlled transformations, but we go beyond them and use additional novel techniques for image discovery and augmentation.

\subsubsection{Visually Similar Image Search}
\label{sec:similar_search}
As the first strategy, we leverage pre-indexed databases that are created to store signatures from millions of images and allow fast retrieval of similar images. The signatures are created from an Inception-v3 based deep learning model trained on all of catalog images for the purpose of product categorization. The embeddings from this model are re-used for various classification and retrieval tasks because they are generic representations of the deep latent factors of the image. In another variation, the signatures are created from VGG16 fc1 layer and then binarized to facilitate efficient indexing. Depending on number of images and the signature size, either FAISS or an ElasticSearch is used for indexing and approximate nearest neighbor search.


\begin{figure}[ht]
\begin{center}
	\centerline{
		\includegraphics[width=\columnwidth]{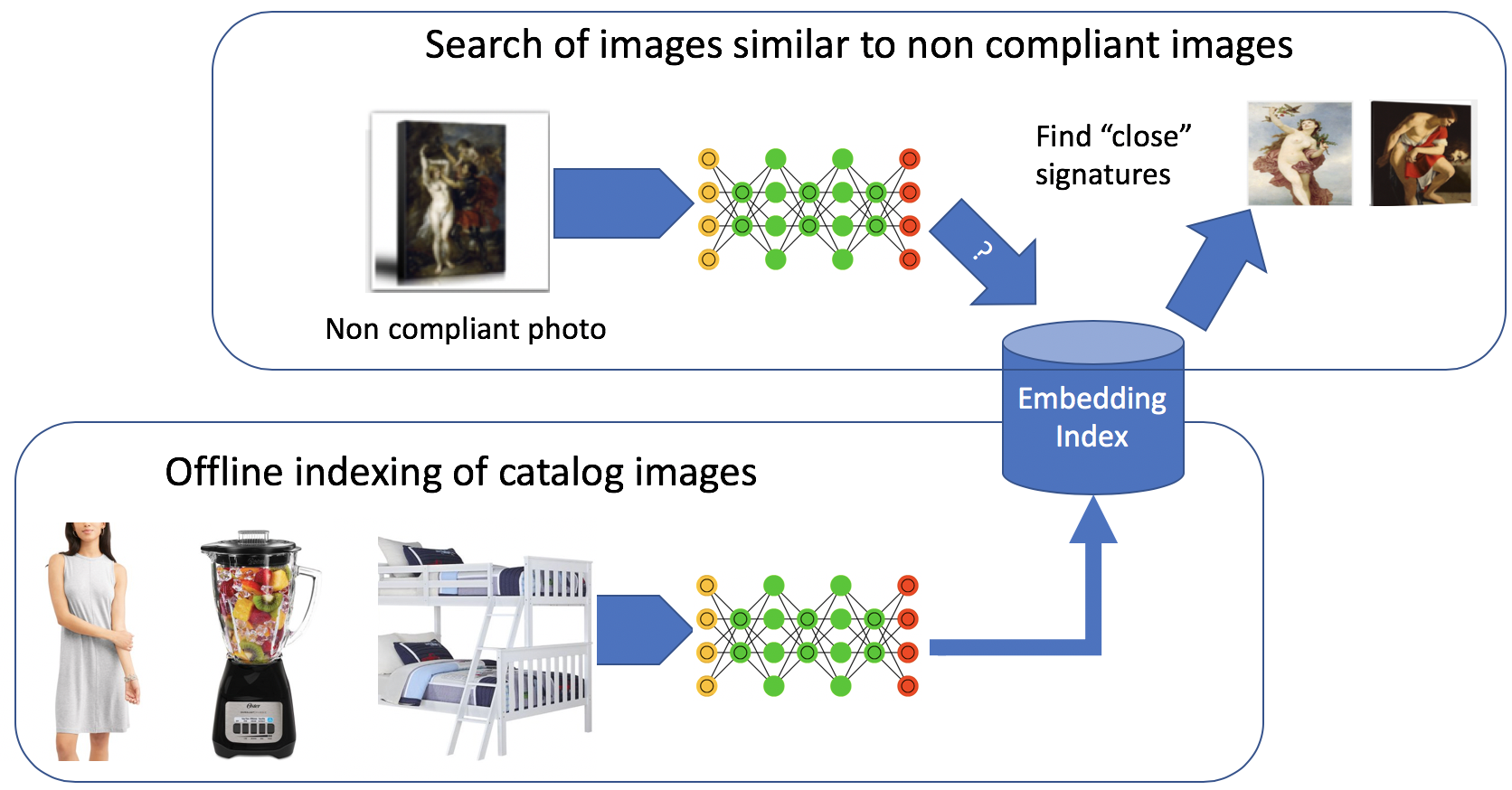}}
		\caption{ Building training data using visual search}
		\label{fig:_Building_training_data_using_visual_search}
\end{center}
\end{figure}


As shown in Figure~\ref{fig:_Building_training_data_using_visual_search}, for every training image, we compute its embedding using either model and then retrieve its nearest neighbors. We manually review the top few results and add some of them to the training set. Similar yet non-offensive images are added as valuable negative training data. For example, search with an image with nudity often retrieves underwear or lingerie model images which are not deemed offensive, but they serve as valuable training data.












\subsubsection{Superimposition of Offensive Content}

The above technique is effective for use cases where the entire object is prohibited such as assault rifle. However, we propose a different method for use cases like logos and badges where the problematic content is a very small part of the product image. Similar image search in such case would not work because the deep learning based signatures have more information about the main object in an image. For example, search by a hat with a certain brand logo will retrieve various hat images, not images of other products with the same brand. 

\begin{figure}[ht]
\vskip -0.1in
\begin{center}
	\centerline{
		\includegraphics[width=\columnwidth]{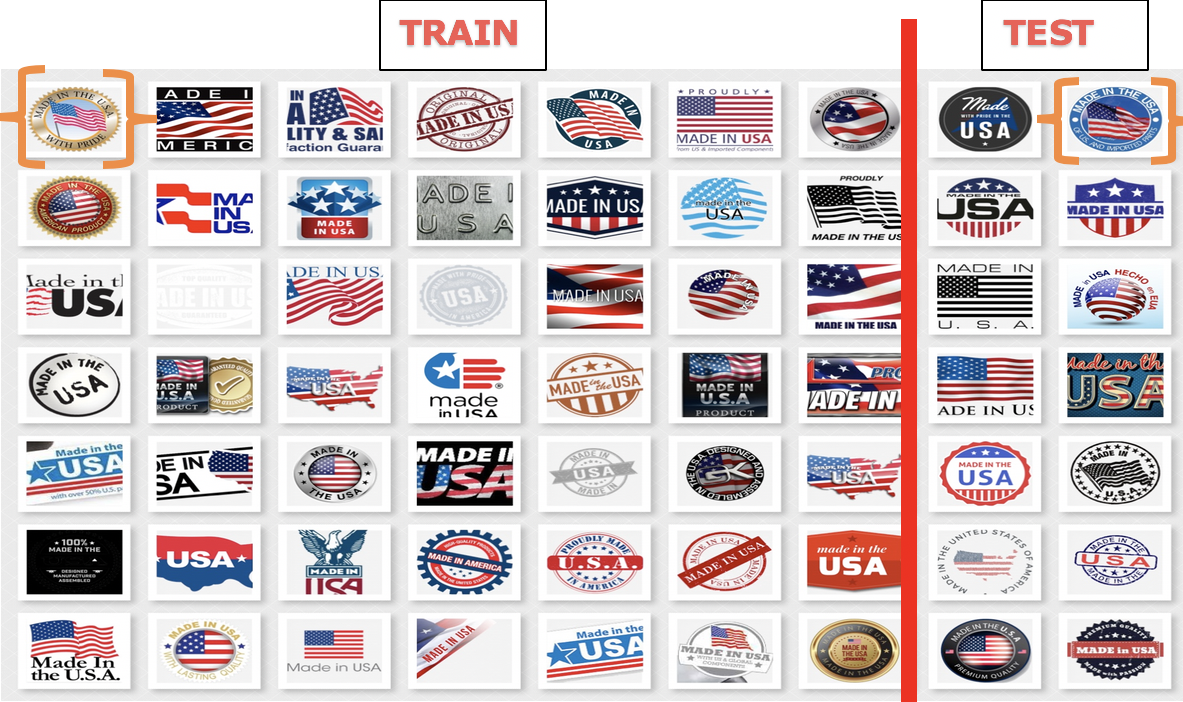}}
		\caption{ Step a $\&$ c: Synthetic data generation using superimposition}
		\label{fig:step_a_b_superimposition}
\end{center}
\vskip -0.3in
\end{figure}

\begin{figure}
\begin{center}
    \centerline{
		\includegraphics[width=\columnwidth]{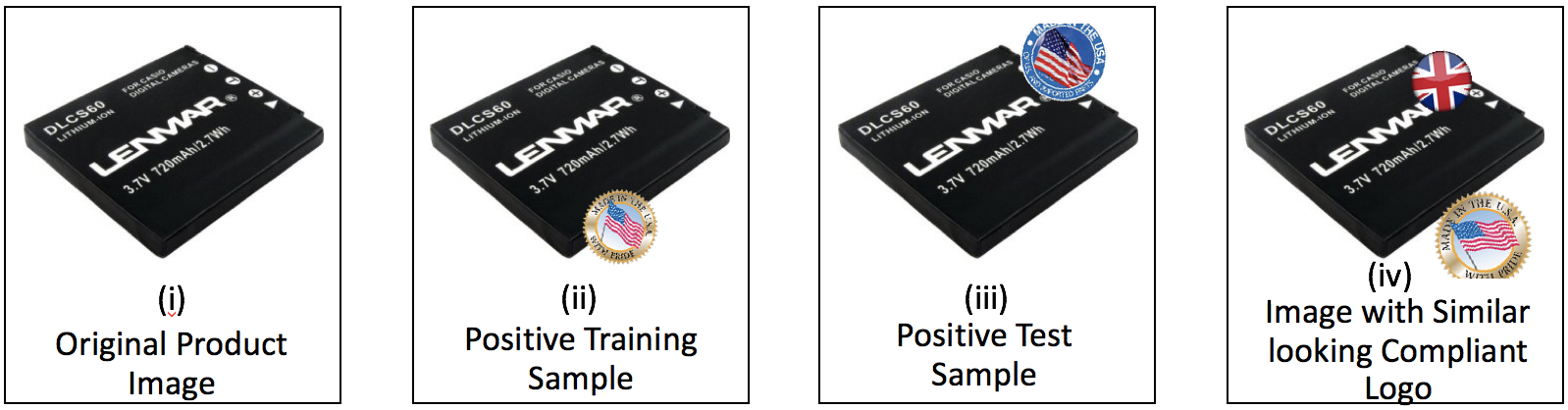}}
		\caption{ Step d: Used training logo (top-left from figure \ref{fig:step_a_b_superimposition}) and applied random scaling, rotation and translation to generate a positive training sample as image (ii). Similarly, testing logo (top-right from figure \ref{fig:step_a_b_superimposition}) used to generate a test sample as image (iii). Image (iv) - Use similar-looking compliant logos for superimposition.}
		\label{fig:step_c_d_e}
\end{center}
\vskip -0.3in
\end{figure}

We address this issue by generating synthetic training images in the following manner:
\begin{enumerate}[label=(\alph*)]
	\item We collect a large number of different-looking logos from the internet or from the data provider. We split the logo images into train and test sets (Figure~\ref{fig:step_a_b_superimposition}). 
	
	\item Not only non-compliant logos, we also collect images of similar-looking compliant logos whenever we have information about them. They will contribute to valuable negative examples. For example, confederate flag and Mississippi flag are quite similar looking.

	\item We tightly crop the logo images, leaving no space around and make the image transparent. (Figure~\ref{fig:step_a_b_superimposition})

	\item We apply controlled transformations on the logos, and then superimpose these logos on regular compliant images to make a non-compliant training or test sample. (Figure~\ref{fig:step_c_d_e} - [ii] and [iii]). Compliant logos are used to create compliant training or test samples (Figure~\ref{fig:step_c_d_e} - [iv]). Transformations include random scaling, rotating, orienting, flipping, translating, mangling, and/ or distorting the non-compliant content. (Figure~\ref{fig:step_c_d_e} - [ii] and [iii]) 
\end{enumerate}

Starting with approximately 100,000 compliant images sampled across the catalog representing all product categories, we apply the above mentioned steps to synthesize 100,000 positive samples for each type of non-compliant logo. Steps (b) and (d) help the model generalize better and reduce false positive rate. Since we know the exact location of superimposition for every image, this process generates a large number of images with logos as well as accurate locations. Obtaining the locations at no cost is a big advantage of this synthetic data generation technique. This dramatically reduces the cost of image annotation, especially when training object detection models.

\subsubsection{Crowdsourcing on Baseline Model Predictions}


\begin{figure}[ht]
\vskip -0.1in
\begin{center}
	\includegraphics[width=\columnwidth]{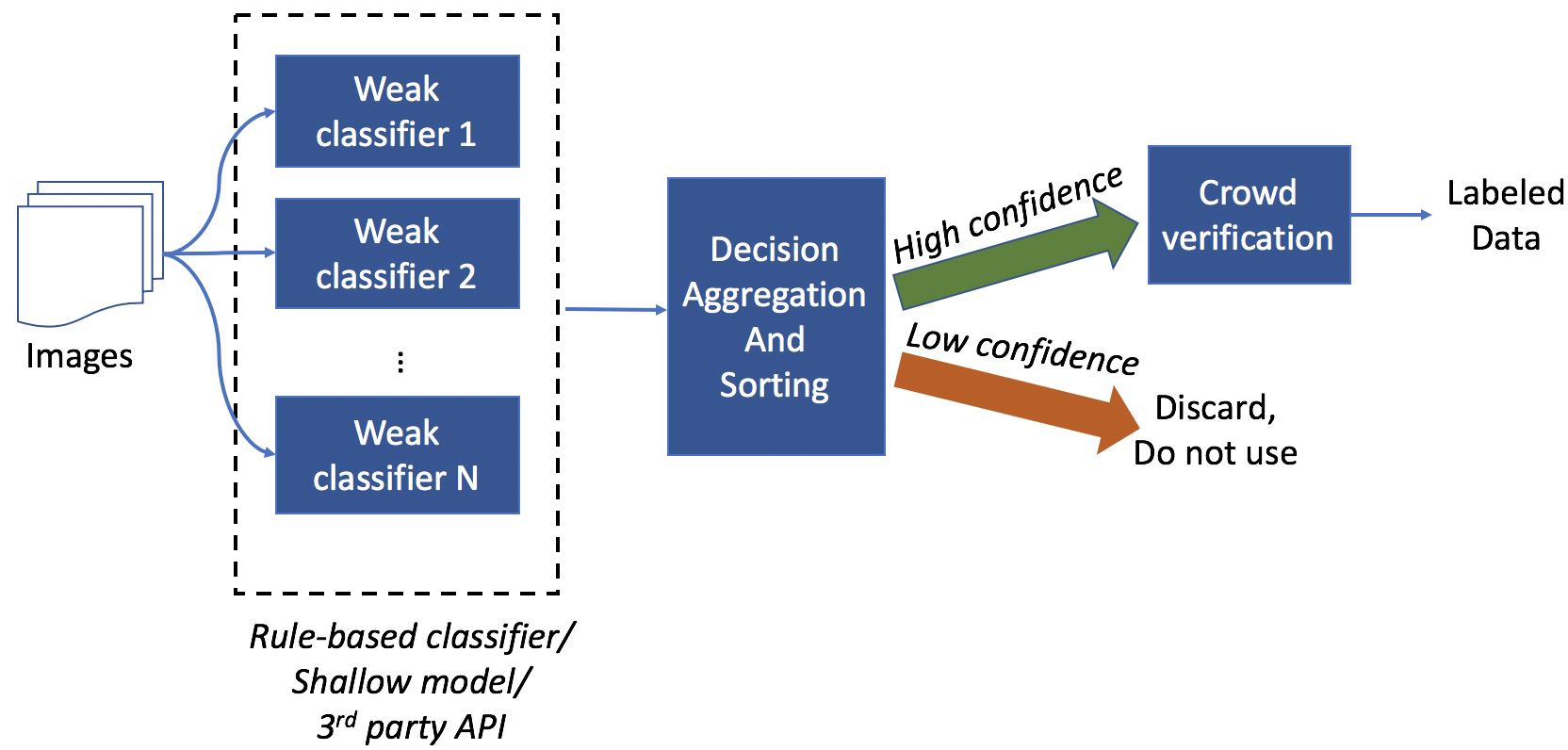}
	\caption{Manual verification of high confidence predictions from baseline predictions create better training data}
	\label{fig:Dealing_with_sparsity}
\end{center}
\end{figure}


We use the training data generated through the above mentioned processes to build shallow linear classifiers, small neural nets and heuristic based classifiers. In some case, we have access to commercially available classifiers from 3$^{rd}$ party. All of these serve as baseline predictors that work as low precision and moderate recall. They are not nearly as good as the required level, however, we use them for a specific purpose (Figure~\ref{fig:Dealing_with_sparsity}). We run them on thousands of images from the catalog to generate predictions with confidence score. Depending on the available crowdsourcing budget, we decide on a confidence threshold. We discard most of the predictions and send only the ones with confidence above that threshold to crowd or trained manual reviewers. They verify the baseline predictions and hence, generate high quality training data. Use of baseline predictors dramatically increases the return on investment for labeling because the high confidence predictions are more likely to be accurate.
 \subsection{Model Training and Prediction}
\label{sec:approaches}
Depending on the amount of training data and the size and shape of the content to be detected, we employ one of the following three approaches.
\subsubsection{Shallow Classifier on Deep Embedding}
\label{subsec:approach1}


\begin{figure}[ht]
\begin{center}
	 \centerline{
		\includegraphics[width=\columnwidth]{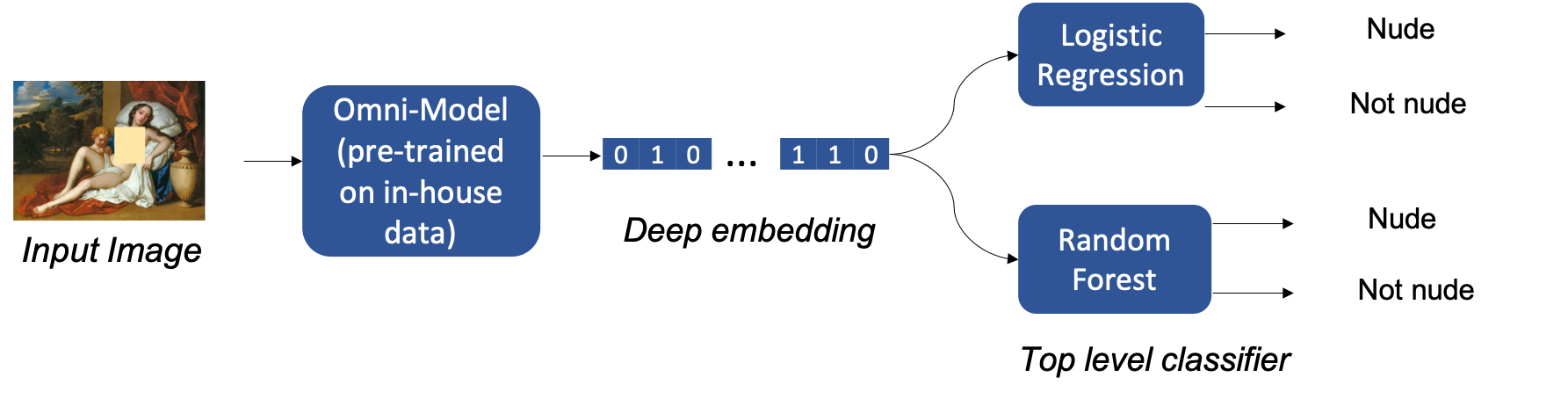}}
		\caption{Shallow Classifier over Deep Embeddings}
		\label{fig:approach1}
\end{center}
\end{figure}


For some problems, we build a shallow classifier on top of image signatures (Figure~\ref{fig:approach1}) extracted from the image similarity models mentioned in Section~\ref{sec:similar_search}. These models are trained on images from the entire product catalog. Logistic Regression and Random Forest are our usual choices for shallow classifier.

\subsubsection{Fine-Tuned Deep Neural Nets}
\label{subsec:approach2}


\begin{figure}
\begin{center}
	 \centerline{
		\includegraphics[width=\columnwidth]{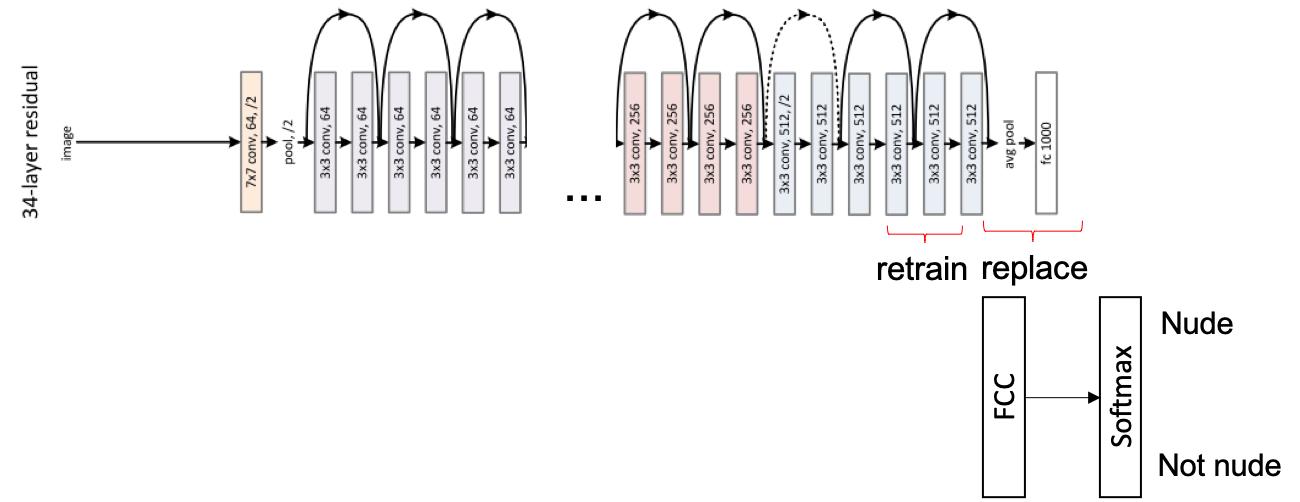}}
		\caption{Approach 2: Retrain Last Layers of Deep CNN}
		\label{fig:approach2}
\end{center}
\end{figure}


For problems where a pre-trained model is less likely to have learnt the concept or the object we need to detect, we retrain the last few layers of that pre-trained network with our data (Figure~\ref{fig:approach2}). We often experiment with Resnet50 and Inception-V3, which were both pretrained on Imagenet dataset. We remove the classification layer and add a fully connected layer and a softmax at the end. We vary the number of residual layers or the inception blocks to be retrained, to find optimal performance. 

\subsubsection{Object Detection}
\label{subsec:approach3}
We use object detection for problems where fine-tuned classification networks do not perform well enough and we have images with annotated bounding boxes. We retrain Faster R-CNN to detect smaller objects such as logos, and we retrain YoloV3 to detect medium to large objects such as frontal nudity, sex toys or assault rifles. For YOLOV3, we run K-means clustering on the annotation boxes in the training data to determine a set of anchor boxes that represent the objects to be detected.

Both YOLOV3 and Faster R-CNN output one or many boxes with labels and confidence scores. We use the most confident boxes to make a decision to block the image or to sent it to human reviewers. Even if they flag the images as compliant, these images contribute as valuable training data.. The object detector output is relatively more explainable than the classification methods because of the boxes.
\begin{figure}[ht]
\begin{center}
	 \centerline{
		\includegraphics[width=\columnwidth]{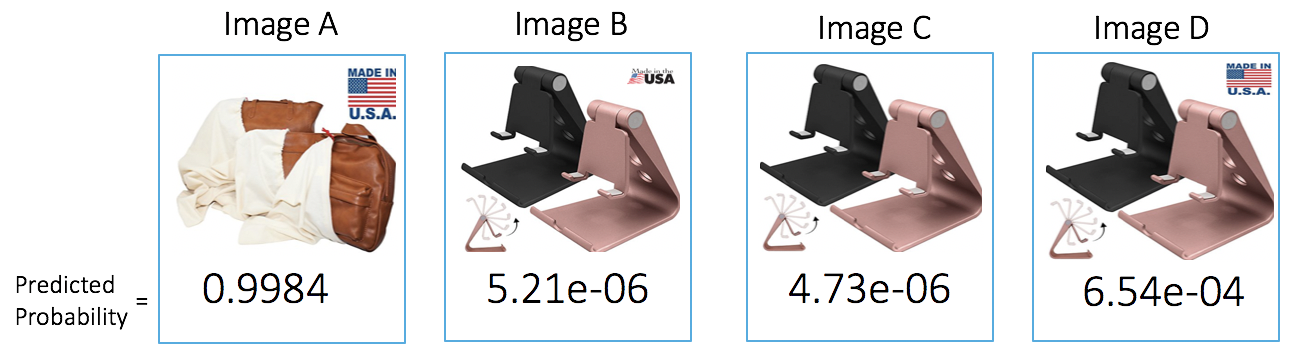}}
		\caption{An example justifying the switch to object detectors from classification models.}
		\label{fig:aha_moment}
\end{center}
\vskip -0.2in
\end{figure}
\subsubsection{Selection of Training Method}
\label{sec:selection}
A mix of intuition and data-driven insight drives the choice of technique for a given problem. To give an example, 
as we wanted to understand why the prediction from fine-tuned deep networks was wrong for a couple of logo test images (Figure~\ref{fig:aha_moment}), we first assumed that model is not generalizing well on different variants of the logo. In the example, image A and B both are non-compliant with different versions of \textit{Made In USA} logo. While the model works perfectly fine on Image A, it was unable to detect Image B.  To test our hypothesis, we created Image C which does not cotain a logo at all, and image D that contains the exact same logo that was detected in A. The classification model could not detect the logo in image D, suggesting that the model was making decisions primarily by recognizing the main object and not the logo. Since it would be prohibitively costly to create a dataset comparable to the product catalog in terms of size and variety, we decided to switch to an object detector for this problem.




Table~\ref{tb:approach_summary} presents which approach resulted in best performance for which problem. The ones tried are marked as Y, the best-performing one is marked with an asterisk.
\begin{table}
\begin{center}
\begin{tabular}{|l|c|c|c|}
\hline
Problem & \makecell{Shallow\\Classifier} & \makecell{Deep\\Classifier} & \makecell{Object\\Detector} \\
\hline\hline
Nudity & Y & Y & Y$^{*}$ \\
Weapons & Y & Y$^{*}$ & N \\
Logo & N & Y & Y$^{*}$\\
\hline
\end{tabular}
\label{tb:approach_summary}
\end{center}
\caption{Approaches tried in different detection problems}
\end{table}

\subsubsection{Two-stage Inference}
\label{sec:ensemble}
%
%
\begin{figure}[ht]
\begin{center}
	\centerline{
		\includegraphics[width=\columnwidth]{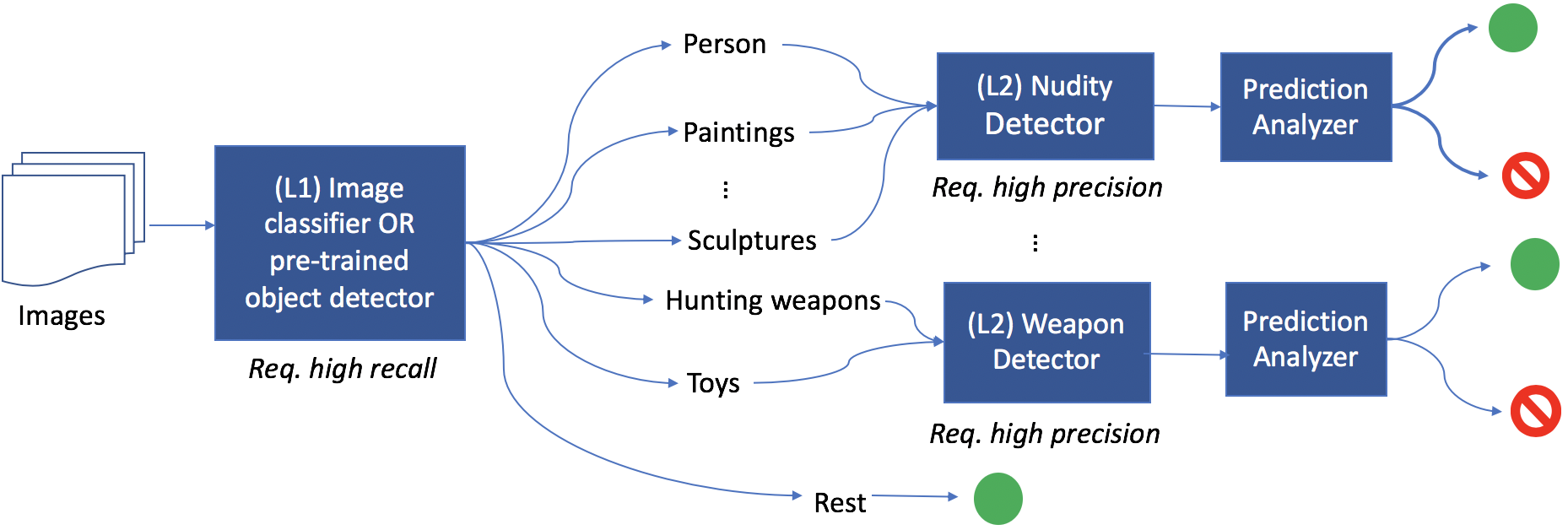}}
		\caption{Two-stage classifiers for non-compliance problems that only occur in selected product categories}
		\label{fig:sys_overview_l1_l2}
\end{center}
\vskip -0.2in
\end{figure}
%
%
In addition to model accuracy, two major driving factors of our system are time and compute resource cost. Running every image of the catalog through an array of deep learning models is prohibitively slow and costly. To address this issue, we make use of the observation that most non-compliance issue is more likely to appear within certain product categories. For example, nudity is most likely to be found in images of people, paintings, sculptures, CDs, carpets, books and posters. Assault rifle images are more likely to be found in hunting gears, toys, and books. This is why we use a broad image classifier as an entry-level filter (Figure~\ref{fig:sys_overview_l1_l2}). This first-level classifier (L1) classifies an input image into one of the major types, such as a person, book, painting etc. Depending on the type of image, it is send to one or more second-level detectors (L2) that are slower in inference and are trained to catch a particular non-compliant category. For example, an image of a person is expected to pass through the nudity detector, an image of a toy is expected to pass through the weapon detector, and so on. If an image does not fall into any of the product types associated with non-compliant categories, it is classified as $`$rest' and it does not go through any L2 detectors.
 \section{System Overview}
\label{sec:system_overview}
\begin{figure}[ht]
\begin{center}
	\centerline{
		\includegraphics[width=\columnwidth]{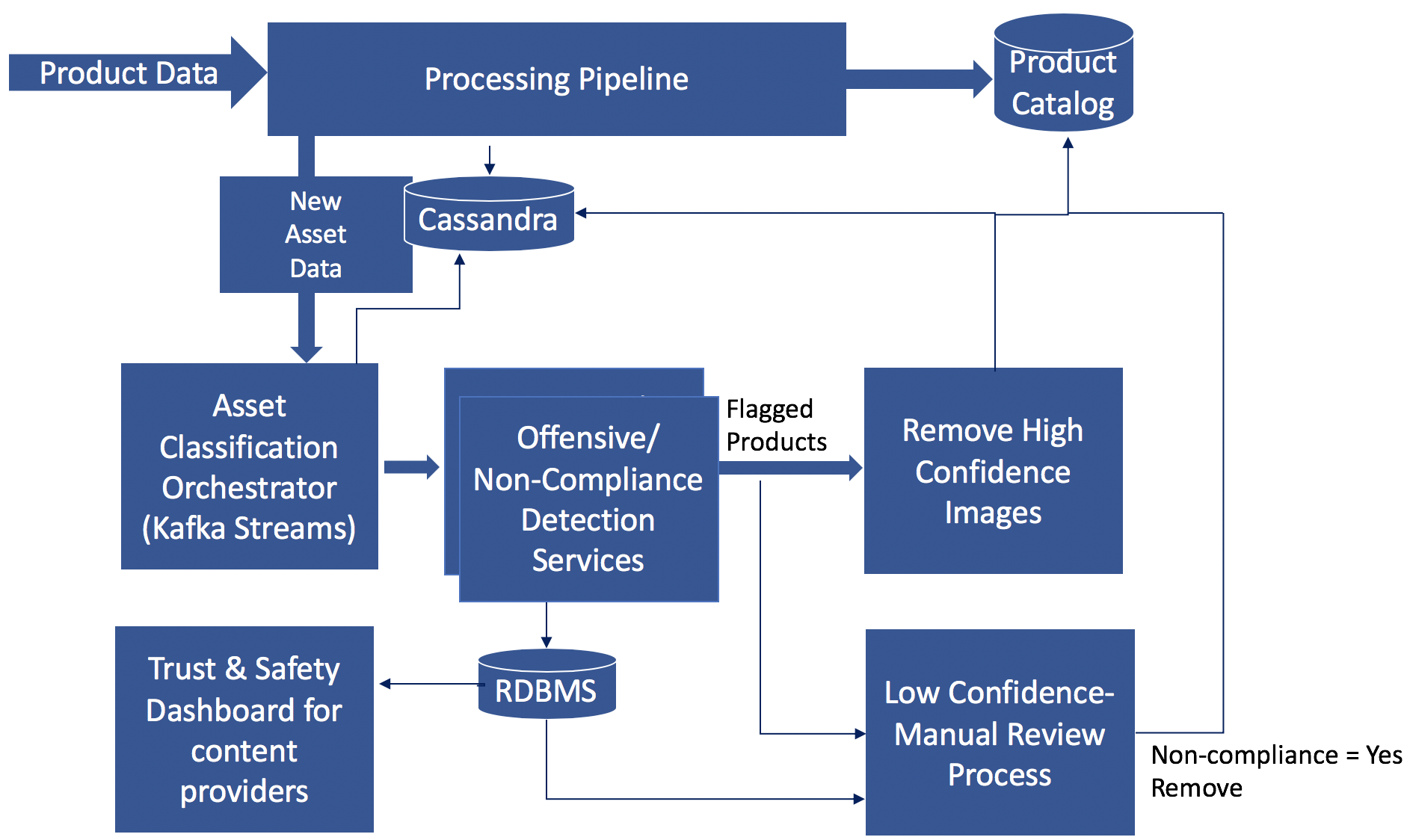}}
		\caption{Overall System architecture}
		\label{fig:Overall_System_architecture}
\end{center}
\vskip -0.3in
\end{figure}

To allow fast and reliable processing of hundreds of thousands of images everyday, the machine learning based detectors (classifiers or object detectors) described above are wrapped into microservices. These microservices are integrated with the the overall image classification engine, as shown in Figure~\ref{fig:Overall_System_architecture}. 


New product data including images is constantly fed to our  e-commerce catalog by suppliers and sellers. The classification engine channels them to a Kafka queue. The queue is read by an orchestrator module that does some pre-processing such as size and format validation. Then, channels the image information to a number of queues dedicated to different detector micro-services. Each micro-service keeps reading from its own queue, processes the image with the model it hosts, and posts the results to a post-processing stream. Images that are flagged as non-compliant with high confidence are automatically removed from the catalog and the corresponding product is blocked. Images flagged with low confidence are pushed to a manual review pipeline. They are either accepted or removed based on manual review response. Sellers and suppliers are given feedback through a dashboard that allows them to review and appeal their blocked content. This image classification system is designed to fit in a bigger product image selection system as in \cite{chaudhuri2018smart}.

 \section{Results}
\label{sec:results}
In this section, we present all the experiments a representative non-compliant category - $``$Best Seller" logos - and an offensive category - nudity. Experiments for other categories are similar in nature and have produced similar results and inferences.

\begin{figure}[ht]
\begin{center}
	\includegraphics[width=\columnwidth]{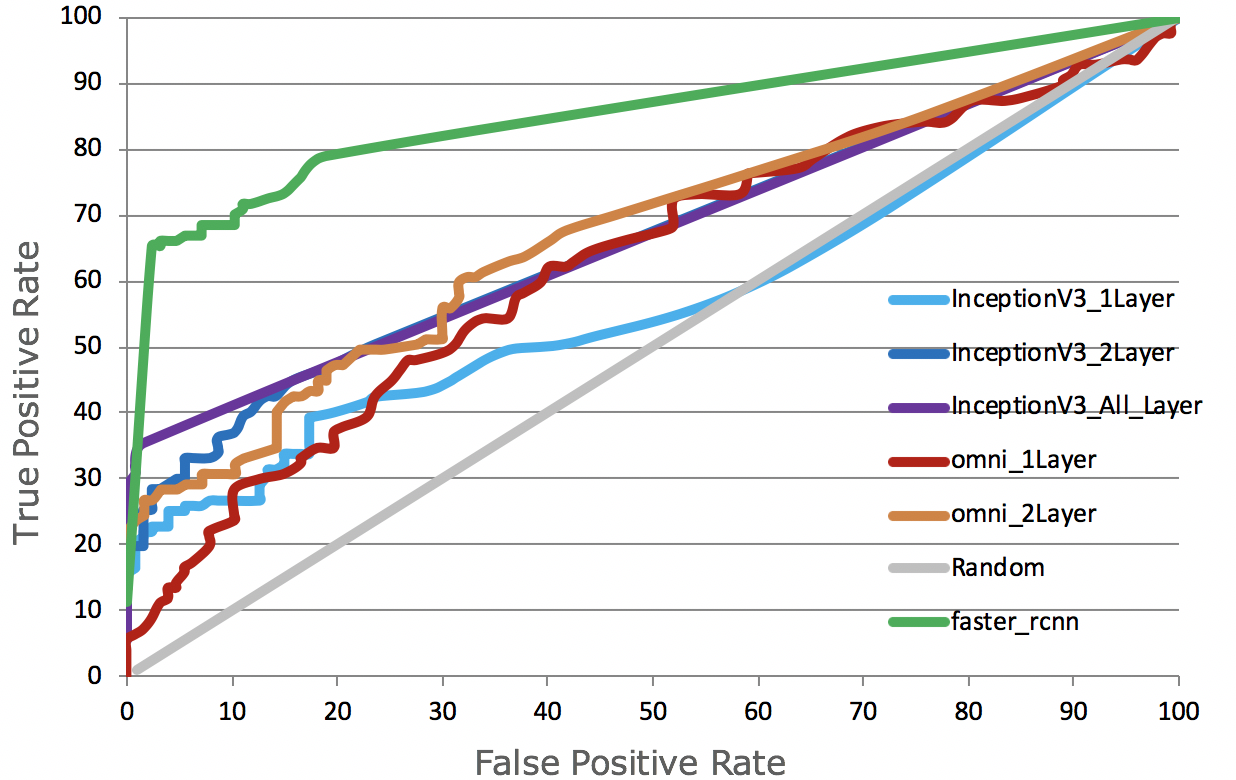}
	\caption{ROC curves based on Approach 2 and 3 for Logo Detection}
	\label{fig:roc_logo}
\end{center}
\vskip -0.2in
\end{figure}

\begin{figure}[ht]
\vskip -0.2in
\begin{center}
	\includegraphics[width=\columnwidth]{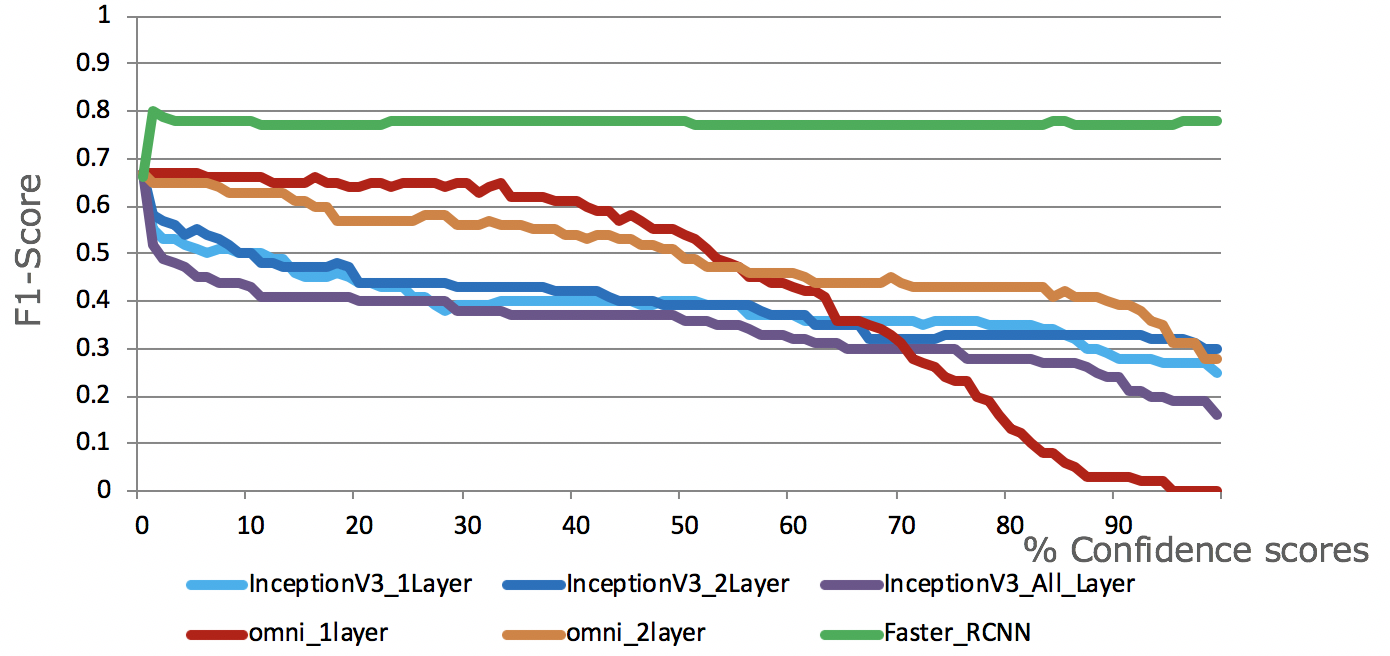}
	\caption{F1-Score for Logo Detection for various confidence thresholds}
	\label{fig:f1_logo}
\end{center}
\vskip -0.2in
\end{figure}

	



	


\begin{table*}[ht]
\caption{Results for Baseline Logo Detector using traditional approaches}
\label{tab:baseline_logo}
\begin{center}
\begin{small}
\begin{sc}
\begin{tabular}{lccc}
\hline
    Technique  & Precision ($\%$ ) & Recall ($\%$ ) & F1-score ($\%$ )\\
\hline
    SIFT + FLANN Matcher & 38.67 &	13.77 &	20.30 \\
    SIFT + BruteForce Matcher & 100.00 &	23.60 &	38.19 \\
    ORB + FLANN Matcher & 49.80	& 8.43 &	14.42 \\
    ORB + BruteForce Matcher & 47.71 &	4.17 &	7.66 \\
    Multi-Scale Template Matching & 45.55 &	4.43 &	8.08 \\
\hline
\end{tabular}
\end{sc}
\end{small}
\end{center}
\vskip -0.2in
\end{table*}

\textbf{Logo and Badge Detection}: We first tried a number of baseline feature matching techniques such as SIFT and ORB feature descriptors followed by  FLANN or BruteForce Matcher.  We also tried multi-scale template matching. The results from these traditional techniques were not satisfactory, as shown in Table~\ref{tab:baseline_logo}. The best f1-score is about 38$\%$. The deep learning techniques performed much better, as shown in Figure~\ref{fig:roc_logo} and \ref{fig:f1_logo}. Linear classification of deep embeddings ((Section~\ref{subsec:approach1}) is not applicable for logos. As for fine-tuned deep nets (Section~\ref{subsec:approach2}), we retrained the last one, two, and all inception layers of a pre-trained InceptionV3. We also experimented with an in-house visual search model which is trained on the entire set of catalog images. We retrained its last one and two layers, the results for which are labelled as \textit{omni\_1layer} and \textit{omni\_2layer} in Figure~\ref{fig:roc_logo} and \ref{fig:f1_logo}. As seen in Figure~\ref{fig:roc_logo}, results from the InceptionV3 and the retrained visual search model are comparable to each other. We also experimented with Faster R-CNN based object detection (Section~\ref{subsec:approach3}). All the experiments were performed on a 700,000 train and 140,000 test set. Figure~\ref{fig:f1_logo} indicates that the f1-score of the Faster R-CNN model is 100$\%$  better than the retrained classification networks at a confidence score of 0.85.

We chose Faster R-CNN since it is known for delivering high accuracy on small objects such as logos. Faster R-CNN is one of the slower models among the popular object detection networks. Since our distributed architecture, designed based on queues allows higher inference time for image analysis, we consciously chose accuracy over inference time.

\begin{figure}[ht]
	\begin{center}
		\includegraphics[width=\columnwidth]{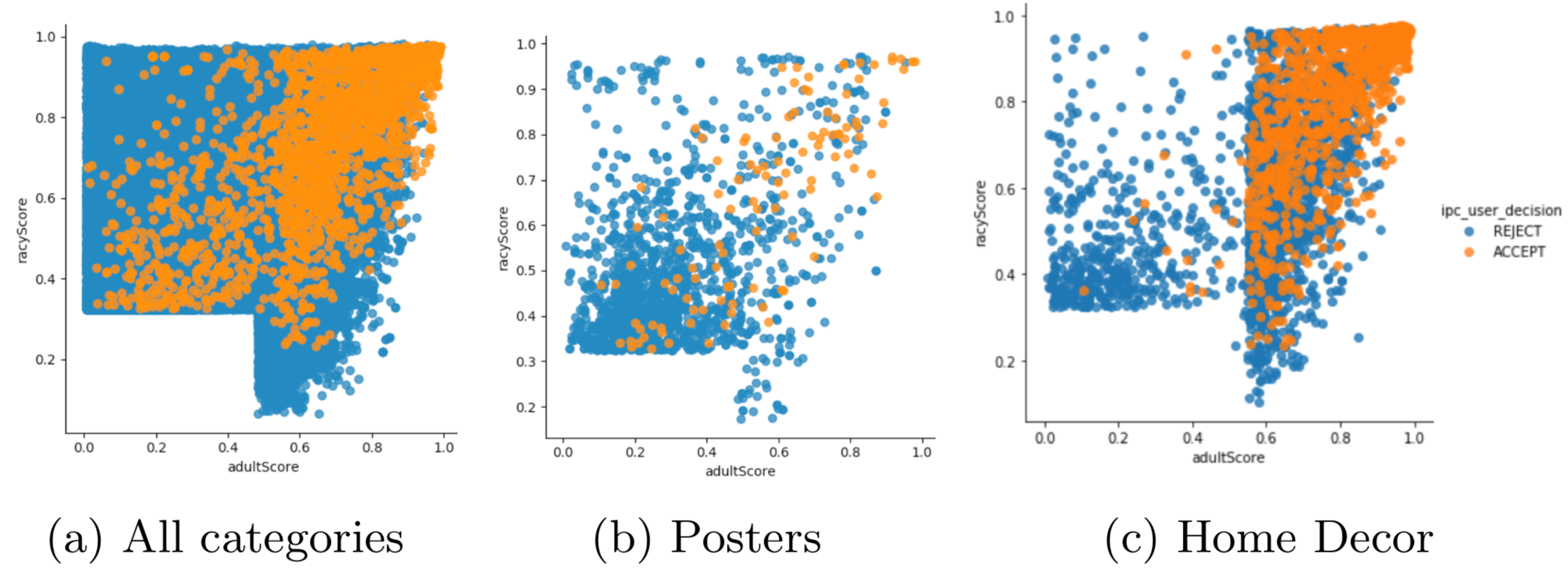}
	\end{center}
	\caption{Manual Review Decision for images flagged by third party API for nudity detection}
	\label{fig:manual_review_for_third_party_API}
\vskip -0.1in
\end{figure}



\begin{table*}[ht]
\caption{Results for Nudity Detector}
\label{tab:results_nudity}
\begin{center}
\begin{small}
\begin{sc}
\begin{tabular}{lccc}
\hline
    Technique  & Precision ($\%$ ) & Recall ($\%$ ) & F1-score ($\%$ )\\
\hline
    \textbf{3rd Party API + Manual Review} & \textbf{x} & \textbf{x} & \textbf{x} \\
    Deep Embedding + Shallow \\ Classifier (approach 1) & +30 & +34.5 & +14 \\
    Resnet50 (approach 2) & +32  & +40 & +35.5  \\
    Inception-V3 (approach 2) & +51 & +49 & +49.5 \\
    \textbf{Object Detection (approach 3)} & \textbf{+55} & \textbf{+67} & \textbf{+54} \\
\hline
\end{tabular}
\end{sc}
\end{small}
\end{center}
\vskip -0.2in
\end{table*}

\textbf{Nudity Detection}: Detecting nudity or sexually explicit content is a widespread problem, so there are many solutions available for purchase. Hence, we started with a commercially available third-party API that accepts an image and returns two scores, namely Adult Score and Racy Score, to quantify its offensiveness. The images with the two aforementioned scores above a certain threshold were sent for manual review to crowd workers. Since the third-party API is trained on different distributions of nude/sexual images compared to those in our catalog, the API returned a large number of false positives. As Figure~\ref{fig:manual_review_for_third_party_API}a suggests, the percentage of images accepted by the crowd (denoted by orange dots) is far less than the count of those rejected by the crowd (denoted by blue dots). The FPR varies across categories (Figures~\ref{fig:manual_review_for_third_party_API}b and ~\ref{fig:manual_review_for_third_party_API}c), but it is on the higher side regardless.

A month-long study of the manual review data revealed that (1) the presence of actual positive instances (nude images) was concentrated in a few segments of the catalog, (2) even within those categories, the distribution of positive and negative instances varied widely. Based on these observations, we fine-tuned the overall threshold and introduced category-specific thresholds. Even with all these changes, the best f1-score we could achieve was very below 25$\%$.


The above baseline method did not perform well, but it helped us lay out strategies to create a near optimal training set for the deep learning approaches. Based on crowd responses for different categories, we built a training set that has enough representation of both positive and negative labels across all categories.





With the carefully crafted training data from baseline method, we experimented with the approaches shown in Table~\ref{tab:results_nudity}. All the experiments are run on internal datasets. The goal of these experiments is to find a method that suits our use case/data, so absolute performance numbers are not presented. Instead, in Table~\ref{tab:results_nudity}, we compare a number of candidate techniques against the baseline method shown in bold with an \textbf{‘x’}. 
The results from deep embedding based linear mode, denoted as \textit{Approach 1} (Section~\ref{subsec:approach1}) and fine-tuned classification networks (Section~\ref{subsec:approach2}), denoted as \textit{Approach 2} are much better than the baseline. Also, fine-tuned Inception v3 performs better than fine-tuned Resnet50. Since Approach 1 use signatures from a model trained on e-commerce catalog images and Approach 2 models are only pre-trained on Imagenet, the former technique generalizes better on new unseen images. Training the base model for Approach 1 is costlier though. Approach 2, which is based on a model trained on Imagenet, can be retrained faster and with less data. In general, we observe that the quality and quantity of the data has a greater impact than the modelling choice. Our final observation: Approach 3 (YOLO v3) outperforms the rest. The manual reviewers annotated about 5,000 images with bounding boxes for the nudity use case. Roughly 40$\%$ of those images serve as positive training labels. With a limited number of images, we retrained YOLO v3 (pretrained on Coco dataset) and tested it on about 8000 test examples to achieve a 54$\%$ lift in f1-score from the baseline.

 \section{Conclusion}
\label{sec:conclusion}
In this paper, we present a computer vision powered system that detects and removes offensive and non-compliant images from an e-commerce catalog containing hundreds of millions of items. In addition to describing the core modeling components of the system, we discuss the technical challenges of building a system at such a scale, namely, lack of adequate training data, an extreme class imbalance and a changing test distribution. We also described the critical refinements made to the data and to the modeling techniques to effectively overcome the challenges. The presented system is already deployed in production and it has processed millions of product images. 

We plan to continue the work towards combining image and textual signals from products to build a more effective model. 
We are also trying to allow the system to detect unforeseen types of non-compliant cases with minimal amount of re-training and fine tuning of existing parameters. 

The strategies adopted and the insights gained can be leveraged by content-serving web-based platforms from other domains as well. 



\bibliographystyle{icml2019}
\bibliography{references}

\begin{thebibliography}{29}
\providecommand{\natexlab}[1]{#1}
\providecommand{\url}[1]{\texttt{#1}}
\expandafter\ifx\csname urlstyle\endcsname\relax
  \providecommand{\doi}[1]{doi: #1}\else
  \providecommand{\doi}{doi: \begingroup \urlstyle{rm}\Url}\fi

\bibitem[Arentz \& Olstad(2004)Arentz and Olstad]{arentz2004classifying}
Arentz, W.~A. and Olstad, B.
\newblock Classifying offensive sites based on image content.
\newblock \emph{Computer Vision and Image Understanding}, 94\penalty0
  (1-3):\penalty0 295--310, 2004.

\bibitem[Bianco et~al.(2017)Bianco, Buzzelli, Mazzini, and
  Schettini]{bianco2017deep}
Bianco, S., Buzzelli, M., Mazzini, D., and Schettini, R.
\newblock Deep learning for logo recognition.
\newblock \emph{Neurocomputing}, 245:\penalty0 23--30, 2017.

\bibitem[Chaudhuri et~al.(2018)Chaudhuri, Messina, Kokkula, Subramanian,
  Krishnan, Gandhi, Magnani, and Kandaswamy]{chaudhuri2018smart}
Chaudhuri, A., Messina, P., Kokkula, S., Subramanian, A., Krishnan, A., Gandhi,
  S., Magnani, A., and Kandaswamy, V.
\newblock A smart system for selection of optimal product images in e-commerce.
\newblock In \emph{2018 IEEE International Conference on Big Data (Big Data)},
  pp.\  1728--1736. IEEE, 2018.

\bibitem[Dai et~al.(2016)Dai, Li, He, and Sun]{dai2016r}
Dai, J., Li, Y., He, K., and Sun, J.
\newblock R-fcn: Object detection via region-based fully convolutional
  networks.
\newblock In \emph{Advances in neural information processing systems}, pp.\
  379--387, 2016.

\bibitem[Deng et~al.(2009)Deng, Dong, Socher, Li, Li, and
  Fei-Fei]{imagenet_cvpr09}
Deng, J., Dong, W., Socher, R., Li, L.-J., Li, K., and Fei-Fei, L.
\newblock {ImageNet: A Large-Scale Hierarchical Image Database}.
\newblock In \emph{CVPR09}, 2009.

\bibitem[Di et~al.(2014)Di, Sundaresan, Piramuthu, and Bhardwaj]{image_ecom_14}
Di, W., Sundaresan, N., Piramuthu, R., and Bhardwaj, A.
\newblock Is a picture really worth a thousand words?: - on the role of images
  in e-commerce.
\newblock In \emph{Proceedings of the 7th ACM International Conference on Web
  Search and Data Mining}, WSDM '14, pp.\  633--642, New York, NY, USA, 2014.
  ACM.
\newblock ISBN 978-1-4503-2351-2.
\newblock \doi{10.1145/2556195.2556226}.

\bibitem[Dosovitskiy et~al.(2015)Dosovitskiy, Fischer, Ilg, Hausser, Hazirbas,
  Golkov, Van Der~Smagt, Cremers, and Brox]{dosovitskiy2015flownet}
Dosovitskiy, A., Fischer, P., Ilg, E., Hausser, P., Hazirbas, C., Golkov, V.,
  Van Der~Smagt, P., Cremers, D., and Brox, T.
\newblock Flownet: Learning optical flow with convolutional networks.
\newblock In \emph{Proceedings of the IEEE international conference on computer
  vision}, pp.\  2758--2766, 2015.

\bibitem[Eggert et~al.(2015)Eggert, Winschel, and Lienhart]{eggert2015benefit}
Eggert, C., Winschel, A., and Lienhart, R.
\newblock On the benefit of synthetic data for company logo detection.
\newblock In \emph{Proceedings of the 23rd ACM international conference on
  Multimedia}, pp.\  1283--1286. ACM, 2015.

\bibitem[Girshick et~al.(2016)Girshick, Donahue, Darrell, and
  Malik]{girshick2016region}
Girshick, R., Donahue, J., Darrell, T., and Malik, J.
\newblock Region-based convolutional networks for accurate object detection and
  segmentation.
\newblock \emph{IEEE transactions on pattern analysis and machine
  intelligence}, 38\penalty0 (1):\penalty0 142--158, 2016.

\bibitem[He et~al.(2016)He, Zhang, Ren, and Sun]{he2016deep}
He, K., Zhang, X., Ren, S., and Sun, J.
\newblock Deep residual learning for image recognition.
\newblock In \emph{Proceedings of the IEEE conference on computer vision and
  pattern recognition}, pp.\  770--778, 2016.

\bibitem[Iandola et~al.(2015)Iandola, Shen, Gao, and
  Keutzer]{iandola2015deeplogo}
Iandola, F.~N., Shen, A., Gao, P., and Keutzer, K.
\newblock Deeplogo: Hitting logo recognition with the deep neural network
  hammer.
\newblock \emph{arXiv preprint arXiv:1510.02131}, 2015.

\bibitem[Joly \& Buisson(2009)Joly and Buisson]{joly2009logo}
Joly, A. and Buisson, O.
\newblock Logo retrieval with a contrario visual query expansion.
\newblock In \emph{Proceedings of the 17th ACM International Conf. on
  Multimedia}, pp.\  581--584, 2009.

\bibitem[Kalantidis et~al.(2011)Kalantidis, Pueyo, Trevisiol, van Zwol, and
  Avrithis]{kalantidis2011scalable}
Kalantidis, Y., Pueyo, L.~G., Trevisiol, M., van Zwol, R., and Avrithis, Y.
\newblock Scalable triangulation-based logo recognition.
\newblock In \emph{Proceedings of the 1st ACM International Conference on
  Multimedia Retrieval}, pp.\ ~20. ACM, 2011.

\bibitem[Krizhevsky et~al.(2012)Krizhevsky, Sutskever, and Hinton]{alexnet}
Krizhevsky, A., Sutskever, I., and Hinton, G.~E.
\newblock Imagenet classification with deep convolutional neural networks.
\newblock In Pereira, F., Burges, C. J.~C., Bottou, L., and Weinberger, K.~Q.
  (eds.), \emph{Advances in Neural Information Processing Systems 25}, pp.\
  1097--1105. Curran Associates, Inc., 2012.

\bibitem[Liu et~al.(2016)Liu, Anguelov, Erhan, Szegedy, Reed, Fu, and
  Berg]{ssd}
Liu, W., Anguelov, D., Erhan, D., Szegedy, C., Reed, S., Fu, C.-Y., and Berg,
  A.~C.
\newblock Ssd: Single shot multibox detector.
\newblock In \emph{European conference on computer vision}, pp.\  21--37.
  Springer, 2016.

\bibitem[Mordvintsev \& K(2013{\natexlab{a}})Mordvintsev and
  K]{feature_matching}
Mordvintsev, A. and K, A.
\newblock Feature matching.
\newblock
  \url{https://opencv-python-tutroals.readthedocs.io/en/latest/py_tutorials/py_feature2d/py_matcher/py_matcher.html},
  2013{\natexlab{a}}.

\bibitem[Mordvintsev \& K(2013{\natexlab{b}})Mordvintsev and
  K]{template_matching}
Mordvintsev, A. and K, A.
\newblock Template matching.
\newblock
  \url{https://opencv-python-tutroals.readthedocs.io/en/latest/py_tutorials/py_imgproc/py_template_matching/py_template_matching.html},
  2013{\natexlab{b}}.

\bibitem[Nielsen(1997)]{reading_numbers}
Nielsen, J.
\newblock How users read on the web.
\newblock \url{https://www.nngroup.com/articles/how-users-read-on-the-web/},
  1997.

\bibitem[Redmon et~al.(2016)Redmon, Divvala, Girshick, and Farhadi]{YOLO}
Redmon, J., Divvala, S., Girshick, R., and Farhadi, A.
\newblock You only look once: Unified, real-time object detection.
\newblock In \emph{Proceedings of the IEEE conference on computer vision and
  pattern recognition}, pp.\  779--788, 2016.

\bibitem[Ren et~al.(2015)Ren, He, Girshick, and Sun]{ren2015faster}
Ren, S., He, K., Girshick, R., and Sun, J.
\newblock Faster r-cnn: Towards real-time object detection with region proposal
  networks.
\newblock In \emph{Advances in neural information processing systems}, pp.\
  91--99, 2015.

\bibitem[Romberg \& Lienhart(2013)Romberg and Lienhart]{romberg2013bundle}
Romberg, S. and Lienhart, R.
\newblock Bundle min-hashing for logo recognition.
\newblock In \emph{Proceedings of the 3rd ACM conference on International
  conference on multimedia retrieval}, pp.\  113--120. ACM, 2013.

\bibitem[Romberg et~al.(2011)Romberg, Pueyo, Lienhart, and
  Van~Zwol]{romberg2011scalable}
Romberg, S., Pueyo, L.~G., Lienhart, R., and Van~Zwol, R.
\newblock Scalable logo recognition in real-world images.
\newblock In \emph{Proceedings of the 1st ACM International Conference on
  Multimedia Retrieval}, pp.\ ~25. ACM, 2011.

\bibitem[Simonyan \& Zisserman(2014)Simonyan and Zisserman]{simonyan2014very}
Simonyan, K. and Zisserman, A.
\newblock Very deep convolutional networks for large-scale image recognition.
\newblock \emph{arXiv preprint arXiv:1409.1556}, 2014.

\bibitem[Su et~al.(2017)Su, Zhu, and Gong]{su2017deep}
Su, H., Zhu, X., and Gong, S.
\newblock Deep learning logo detection with data expansion by synthesising
  context.
\newblock In \emph{2017 IEEE Winter Conference on Applications of Computer
  Vision (WACV)}, pp.\  530--539. IEEE, 2017.

\bibitem[Szegedy et~al.(2014)Szegedy, Liu, Jia, Sermanet, Reed, Anguelov,
  Erhan, Vanhoucke, and Rabinovich]{inceptionv1}
Szegedy, C., Liu, W., Jia, Y., Sermanet, P., Reed, S.~E., Anguelov, D., Erhan,
  D., Vanhoucke, V., and Rabinovich, A.
\newblock Going deeper with convolutions.
\newblock \emph{CoRR}, abs/1409.4842, 2014.

\bibitem[Szegedy et~al.(2015)Szegedy, Vanhoucke, Ioffe, Shlens, and
  Wojna]{inceptionv2}
Szegedy, C., Vanhoucke, V., Ioffe, S., Shlens, J., and Wojna, Z.
\newblock Rethinking the inception architecture for computer vision.
\newblock \emph{CoRR}, abs/1512.00567, 2015.

\bibitem[Zakrewsky et~al.(2016)Zakrewsky, Aryafar, and
  Shokoufandeh]{qualitypopularity}
Zakrewsky, S., Aryafar, K., and Shokoufandeh, A.
\newblock Item popularity prediction in e-commerce using image quality feature
  vectors.
\newblock \emph{CoRR}, abs/1605.03663, 2016.

\bibitem[Zheng et~al.(2004)Zheng, Liu, and Daoudi]{zheng2004blocking}
Zheng, H., Liu, H., and Daoudi, M.
\newblock Blocking objectionable images: adult images and harmful symbols.
\newblock In \emph{2004 IEEE International Conference on Multimedia and Expo
  (ICME)(IEEE Cat. No. 04TH8763)}, volume~2, pp.\  1223--1226. IEEE, 2004.

\bibitem[Zoph et~al.(2017)Zoph, Vasudevan, Shlens, and Le]{nasnet}
Zoph, B., Vasudevan, V., Shlens, J., and Le, Q.~V.
\newblock Learning transferable architectures for scalable image recognition.
\newblock \emph{CoRR}, abs/1707.07012, 2017.

\end{thebibliography}

\end{document}